\documentclass{ecai} 

\usepackage{latexsym}
\usepackage{amssymb}
\usepackage{amsmath}
\usepackage{amsthm}
\usepackage{booktabs}
\usepackage{enumitem}
\usepackage{graphicx}
\usepackage{color}

\newcommand{\BibTeX}{B\kern-.05em{\sc i\kern-.025em b}\kern-.08em\TeX}

\usepackage{listings}
\usepackage{xurl}
\usepackage{hyperref}
\usepackage{breqn}
\usepackage{ragged2e}
\usepackage[htt]{hyphenat}

\usepackage{algorithm}
\usepackage{algpseudocode}
\usepackage{array}
\makeatletter
\newcommand{\thickhline}{%
    \noalign {\ifnum 0=`}\fi \hrule height 1.5pt
    \futurelet \reserved@a \@xhline
}
\newcolumntype{"}{@{\hskip\tabcolsep\vrule width 1.5pt\hskip\tabcolsep}}
\newcommand{\thickhlineSecond}{%
    \noalign {\ifnum 0=`}\fi \hrule height 1pt
    \futurelet \reserved@a \@xhline
}

\definecolor{codegreen}{rgb}{0,0.6,0}
\definecolor{codegray}{rgb}{0.5,0.5,0.5}
\definecolor{codepurple}{rgb}{0.58,0,0.82}
\definecolor{backcolour}{rgb}{0.95,0.95,0.92}

\lstdefinestyle{mystyle}{
    backgroundcolor=\color{backcolour},   
    commentstyle=\color{codegreen},
    keywordstyle=\color{magenta},
    numberstyle=\tiny\color{codegray},
    stringstyle=\color{codepurple},
    basicstyle=\ttfamily\footnotesize,
    breakatwhitespace=false,         
    breaklines=true,                 
    keepspaces=true,                 
    numbers=left,                    
    numbersep=5pt,                  
    showspaces=false,                
    showstringspaces=false,
    showtabs=false,                  
    tabsize=2
}

\begin{document}


\begin{frontmatter}


\paperid{123} 


\title{A Fuzzy Approach to the Specification, Verification and Validation of Risk-Based Ethical Decision Making Models}


\author[A]{\fnms{Abeer}~\snm{Dyoub}\orcid{0000-0003-0329-2419}\thanks{Corresponding Author. Email: abeer.dyoub@uniba.it.}}
\author[A,B]{\fnms{Francesca A.}~\snm{Lisi}\orcid{0000-0001-5414-5844}}

\address[A]{{Dept. of Informatics, University of Bari ``Aldo Moro'', Bari, Italy}}
\address[B]{Centro Interdipartimentale di Logica e Applicazioni (CILA), University of Bari ``Aldo Moro'', Bari, Italy}


\begin{abstract}
The ontological and epistemic complexities inherent in the moral domain make it challenging to establish clear standards for evaluating the performance of a moral machine. In this paper, we present a formal method to describe Ethical Decision Making models based on ethical risk assessment. Then, we show how these models that are specified as fuzzy rules can be verified and validated using fuzzy Petri nets. A case study from the medical field is considered to illustrate the proposed approach.
\end{abstract}

\end{frontmatter}


\section{Introduction}
\label{intro}

\textit{Ethical Decision Making} (EDM) is a multifaceted process that involves numerous factors that blend both reasoning and emotions. It is highly flexible, context-dependent, and varies across cultures. Since the early 21st century, several attempts have been made to integrate EDM into AI-based systems. However, no universally accepted or fully reliable model has yet been developed.
Most of the existing approaches to EDM are based on ethical theories from moral philosophy which belong mainly to one of the three primary categories of ethics schools, namely Consequentialism, Deontology, and Virtue Ethics \cite{Frischhut2019}.
These ethical theories are fundamentally incompatible with each other. 
They often provide different moral answers to the same dilemma. Thus, the idea of solving ethical issues by combining elements of Utilitarianism, Kantianism, and perhaps Virtue Ethics is a flawed approach to creating ethically sound AI \cite{blackman2022ethical}. These theories primarily aim to explain what is right or wrong and why, rather than serving as frameworks for EDM procedures. However, ethical reasoning is much more complex than adhering to any singular moral philosophical theory.

Unprecedented challenges to EDM have arisen with the advent of the so-called \textit{Symbiotic Artificial Intelligence} (SAI). 
It represents a shift from traditional AI by fostering a collaborative, mutually beneficial relationship between humans and AI. Rather than replacing human abilities, SAI augments and adapts to them. As the relationship between humans and AI deepens, AI actions can have a greater impact on human lives. Therefore, with SAI there is an increasing need for machine ethics to maximize the benefit and mitigate the risk of causing harm to the society arising from SAI systems. This is essential to protect human rights, foster trust and ensure long-term human-AI collaboration.
With SAI systems, as the level of symbiosis increases, so does the ethical risk. To ensure that these systems behave ethically and do not cause harm of any kind, we believe that when building EDM for SAI systems, we should take the following steps. First, we need to consider the domain in which these SAI systems will operate. Second, we need to identify possible ethical risks, i.e. risks of causing harm which could be of different types: physical, mental, violation of privacy, discrimination or any other violation of human rights or values. Third, based on the level of ethical risk, we need to choose the best action/decision to mitigate that risk. So, Ethical Risk Assessment (ERA) appears to be a crucial phase in the EDM process for the case of SAI systems and poses several issues.

A major problem is the difficulty of accurately estimating the possible ethical risks without a complete understanding of all aspects of the risk system being studied. In practical scenarios, it is impossible to completely eliminate gaps in ERA, resulting in fuzziness (imprecision, vagueness, incompleteness, etc.) that we need to address and manage. Moral decisions are indeed not always black and white but exist on a spectrum of ethical considerations, much as fuzzy logic deals with degrees of truth rather than absolute values. Morality is therefore a fuzzy concept because it lacks clear boundaries and varies according to context, culture, and individual perspective. This makes it amenable to fuzzy modeling.

Another open issue is the need of EDM models that can be formally evaluated.
Most existing works on the implementation of machine ethics provide no evaluation or only an informal evaluation of the proposed models \cite{TolmeijerKSCB21}. Some works are illustrated through a few example scenarios or exhibit application domains, which leads to limited validity. Some others use the test approach to compare the outcomes of their systems with a ground truth. However, very few of them validate their claims with some formal proof. 

In this work, we first propose a fuzzy approach for ERA of SAI systems. Then, we propose to represent the overall EDM model by a set of fuzzy rules. 
The ERA system constitutes one module in the overall EDM system.
We give a formal definition of the EDM model. This definition will allow us to clearly describe the EDM model and construct validation referents. Furthermore, it allows us to map the model to a Petri net, which subsequently allows us to apply formal verification and validation (V\&V) criteria to these models (\cite{814327,1185845}).
The paper is structured as follows: related works are discussed in Section \ref{related}. In Section \ref{pre} we give some background on Fuzzy Logic and Fuzzy Petri Nets. We present our proposal for EDM modeling in Section \ref{proposed}, dedicating the Subsection \ref{era} to our proposed approach for ERA, and the Subsections \ref{ver} and \ref{val} to V\&V. Then, in Section \ref{case}, we present a case study in the medical domain to illustrate our proposal. Finally, we conclude with Section \ref{con}.

\section{Related Works}
\label{related}

In this section, we are concerned only with the works that provide, in addition to the proposal for building the ethical machine, a formal evaluation of their machine according to the principles of V\&V.

Dennis \textit{et al.} \cite{DennisFW15} build on the consequence engine proposed by \cite{WinfieldB014}. This consequence engine is a discrete component of an
autonomous system that integrates together with methods for action selection in the robotic controller. It evaluates the outcomes of actions using simulation and prediction, and selects the most ethical option using a safety/ethical logic. The authors develop a declarative language to specify such consequence engines as agents implemented within the agent infrastructure layer toolkit (AIL). The systems developed using AIL are verifiable in the AJPF model checker \cite{DennisFWB12}. The authors reimplement a version of the case study reported in \cite{WinfieldB014}  as an agent and show how the operation of the consequence engine can be verified with AJPF. 
Murakami \cite{Murakami04} presents an axiomatization of Horty’s utilitarian formulation of multi-agent deontic logic. Conversely, Arkoudas et al. \cite{BringsjordAB06} propose a subordinate-based deduction formulation of Murakami's system; then they implement this system and prove it using the Athena theorem prover. 
Anderson et al. \cite{AndersonAB19} advocate an ethical Turing test, to validate the behavior of their ethical agent, where the ethically preferable action specified by a machine faced with an ethical dilemma is compared with the choice that an ethicist would take when faced with the same
ethical dilemma. If a significant number of answers given by the machine match the answers given by the ethicist, then it has passed the test.

One of the advantages of our proposal for dealing with EDM models is that it allows us to exploit a well-defined, formal, and mathematically rigorous structure like Petri Nets to specify, verify, and validate EDM models.

\section{Preliminaries}
\label{pre}
\subsection{Fuzzy Logic in a Nutshell}
\label{fuzz}
Developed by Lotfi Zadeh in the 1960s, fuzzy logic \cite{zadeh1988fuzzy} is based on fuzzy set theory, that is a generalization of the classical set theory. The classical sets are also
called clear sets, as opposed to vague, and similarly classical logic is also
known as Boolean logic or binary logic. 
A \textit{fuzzy set} is a mathematical construct that allows an element to have a gradual degree of membership within the set, as opposed to the binary inclusion found in classical sets \cite{zimmermann2011fuzzy}. Formally, a fuzzy set \( A \) in a universe of discourse \( X \) is defined by a \textit{membership function} \( \mu_A : X \rightarrow [0, 1] \), where each element \( x \in X \) is assigned a degree of membership \( \mu_A(x) \). This value represents the extent to which \( x \) belongs to the fuzzy set \( A \). Membership functions (MF) can take various shapes, such as triangular, trapezoidal, or Gaussian, depending on the problem domain and the nature of the input data \cite{ross2005fuzzy}.
In fuzzy logic, reasoning, also known as \textit{approximate reasoning}, is based on fuzzy rules that can be easily expressed in natural language using linguistic variables such as "HIGH" or "LOW" like in:\\
\textit{If (the quality of the food is HIGH), then (the tip is HIGH)}.\\ 
Formally, a \textit{fuzzy rule} has the form:
\[
\text{If } x \in A \text{ and } y \in B \text{, then } z \in C,
\]
where \( A \), \( B \), and \( C \) are fuzzy sets.
For more details on fuzzy logic and its applications, the reader can refer, among many, to \cite{zimmermann2011fuzzy,tamir2015fifty,rea2022risk}.

\subsection{Fuzzy Petri Nets: a Brief Introduction}
\label{petri}
A fuzzy Petri net (FPN) \cite{1715490,60794} is a directed graph
containing two types of nodes: places and transitions, where
circles represent places and bars represent transitions. FPN can be used to represent fuzzy rule-based systems. Each
place represents an antecedent or consequent and may or may
not contain a token associated with a truth degree between
zero and one which speaks for the amount of trust in the
validity of the antecedent or consequent. Each transition
representing a rule is associated with a certainty factor value
between zero and one. The certainty factor represents the
strength of the belief in the rule. The relationships from places
to transitions and vice versa are represented by directed arcs.
The concept of FPN is derived from Petri nets. FPN structure can be defined as an 8-tuple:
\begin{equation}
N = (P, T,D, I, O, f, \alpha, \gamma)
\end{equation}
where each component is defined as follows:\\
\begin{math}
P = \{ p_1, p_2, ... , p_n \} \text{ is a finite set of places};\\
T = \{ t_1, t_2, ... , t_m\} \text{ is a finite set of transitions};\\
D = \{ d_1, d_2, ... , d_n \} \text{ is a finite set of propositions}; \\
I : P \times T \rightarrow \{0,1\} \text{ is an input function, from places to transitions};\\
O : T \times P \rightarrow \{0,1\}\text{ is an output function, from
transitions to places};\\
f : T \rightarrow [0,1] \text{ is an association function, from transitions to [0,1]};\\
\alpha : P \rightarrow [0,1] \text{ is an association function, from places to [0,1]};\\
\gamma : P \rightarrow D \text{ is an association function, from places to propositions}.
\end{math}

Note that $P \cap T \cap D = \emptyset$ and $|P| = |D|$.

\subsection{Mapping Rule-Bases to FPN}

In order to map our rule-base to FPNs, we follow the steps introduced in \cite{1185845} by first, normalizing the rules, and second transforming the normalized rules into Petri nets.

\paragraph{Rules Normalization}
This step is preliminary to the transformation of the rules into their corresponding Petri nets. Our interest is in propositional clauses that allows only on consequent part in each rule. Normalization leverages logical equivalences to transform complex logical expressions into simpler, more readable forms \cite{buning1999propositional,robinson2001handbook}. 
A propositional logic clause of the form $P_1 \wedge P_2 \wedge ... \wedge P_{j-1} \rightarrow P_j$, where $ P_i$'s are propositions, and \( \rightarrow \) and \( \land \) are logical connectives representing implication and conjunction respectively, is logically equivalent to: $\neg(P_1 \wedge P_2 \wedge ... \wedge P_{j-1}) \vee P_j$. 
Based on logical equivalence, rules of these two forms: 1) $P_1 \wedge P_2 \wedge ... \wedge P_{j-1} \rightarrow P_j \wedge P_{j+1} \wedge ... \wedge P_k$; 2) $P_1 \vee P_2 \vee ... \vee P_{j-1} \rightarrow P_j \wedge P_{j+1} \wedge ... \wedge P_k$; can be normalized into the following three rule types:
\begin{itemize}
	\item Type 1: $P_1 \wedge P_2 \wedge ... \wedge P_{j-1} \rightarrow P_i$, where $1 < j \geq i\leq k$.	
    \item Type 2: $P_i \rightarrow P_1 \wedge P_2 \wedge ... \wedge P_{j-1} $, where $1\geq i\leq j-1$. This rule can be divided into a set of rules: $P_i \rightarrow P_1$, $P_i \rightarrow P_2$, ..., $P_i \rightarrow P_{j-1}$.
	\item Type 3: $(P_1 \vee P_2 \vee ... \vee P_{j-1})\rightarrow P_j$, where $1 < j \geq i\leq k$.
\end{itemize}

Certainty factor is a measure of confidence or belief that quantifies how certain we are about the rule’s conclusion based on the conditions \cite{60794}.
Let $\alpha_i$ denote the degree of truth of antecedent / consequent parts $P_i$ of a rule $r_i$ and $\beta_i$ denotes the degree of confidence of the rule $r_i$. We can obtain the rules with certainty factors as follows: 
\begin{itemize}
	\item Type 1: $R_{i}(\beta_i): P_1 (\alpha_1) \wedge P_2 (\alpha_2)\wedge ... \wedge P_{j-1} (\alpha_{j-1})\rightarrow P_j (\alpha_j)\wedge P_{j+1} (\alpha_{j+1})\wedge ... \wedge P_k (\alpha_k)$
 
	\item Type 2: $R_1(\beta_1) : P_j(\alpha_j) \rightarrow P_1(\alpha_1); \quad
R_2(\beta_2) : P_j(\alpha_j) \rightarrow P_2(\alpha_2); \quad \dots \quad;
R_j(\beta_{j-1}) : P_j(\alpha_j) \rightarrow P_{j-1}(\alpha_{j-1}).$

	\item Type 3: $R_{i}(\beta_i): (P_1 (\alpha_1)\vee P_2 (\alpha_2)\vee ... \vee P_{j-1} (\alpha_{j-1}))\rightarrow P_j (\alpha_j)$
\end{itemize}

\paragraph{Transforming rules to FPN}
To represent the EDM model by FPN, we need to map antecedents of a rule $A(F)$ to input places of the FPN, consequents of a rule $C(G)$ to output places of the FPN, and a rule name to a transition of the FPN etc. These mappings can be formally described as shown below:

\noindent
\begin{math}
(1) A(F) \text{ is mapped to} -p_i, \text{where} -p_i  \text{ is an input place;} \\
(2) C(G) \text{ is mapped to } p_i -, \text{ where } p_i - \text{ is an ouput place;}\\
(3) Rule.Rname \text{ is mapped to } t_i, \text{ where } t_i  \text{ is a transition}; \\
(4) Rule.CF \text{ is mapped to } f(t_i),\text{ where } CF \text{ is certainty factor} ;\\
(5) F \text{ is mapped to } \alpha(-p_i), G \text{ is mapped to } \alpha(p_i -);\\
(6) A \text{ is mapped to } \gamma(-p_i), C \text{ is mapped to } \gamma(p_i -).
\end{math}


\section{Our Proposed EDM Model }
\label{proposed}

We propose a fuzzy approach for specification, verification and validation of EDM models. This approach is inspired by research from the field of human behavior modeling \cite{harmon2006validation,4809789,4675474}. Our approach closely models human ethical perception by utilizing fuzzy logic, as opposed to the rigid binary logic of allowed and forbidden actions.
We aim to design a model capable of implementing context-specific ethical principles for a machine, irrespective of its underlying ontological framework, whether driven by utilitarianism, deontology, Kantianism, prima facie duties, cultural norms, religious beliefs, or legal considerations.
Ethical guidelines used to define ethically compliant machine behavior within a specific domain or context can be developed through consultation with ethicists, legal experts, philosophers, social scientists, technologists, and beneficiaries of the technology, among others. Focusing solely on domain-specific values to ensure ethically compliant machine behavior may be more practical than attempting to establish principles for a universal moral theory.

In our proposed model, EDM is mainly based on the level of ethical risk calculated by a specific module fERA of the fEDM system.
fERA module is also based on fuzzy logic (see Section \ref{era}).
Various input and output fuzzy variables that dictate the required ethical behaviour of the machine are identified. Then, the required ethical behaviour of the machine is expressed in terms of fuzzy rules for EDM. These rules map the inputs, and risk levels to outputs (actions/decisions) of the system (FERDs (see Definition 1)).

\subsection{Fuzzy Rule-Based EDM Model}
\label{fuzzyEDM}

\begin{figure}[htbp!]
  \centering
  \includegraphics[width=0.8\linewidth]{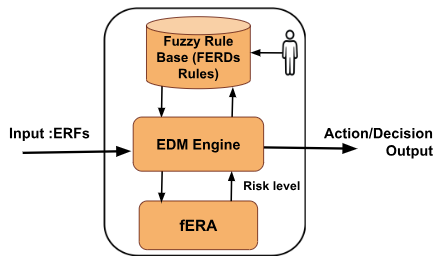}
  \caption{General Structure of the Fuzzy EDM (fEDM) Model }
  \label{fig:EDM}
\end{figure}

 \vspace{15pt}

Figure \ref{fig:EDM} shows the main components of the proposed fEDM model which are described in the following:
\begin{itemize}
    \item The fERA module receives from the EDM engine information about the case at hand, particularly the relevant factors/parameters for ethical risk assessment. fERA uses this information to  calculate the risk level value.
    \item The EDM Engine, at minimum, performs the following functions:
    \begin{itemize}
        \item accepts input about the ethically relevant factors (ERFs) of the case at hand; passes them to the fERA module and receives from fERA the risk level.
        \item consults the Fuzzy Rule Base (FERDs) to generate its responses (actions/decisions).
        \item can learn fuzzy decision rules from data (if available), and add them to the Fuzzy Rule Base to be used later for decision making (in the current implemented version, fuzzy inference rules are written manually by human experts).
    \end{itemize}
    \item The Fuzzy Rule Base contains a set of fuzzy rules that define the behavior of the system. These rules describe how fuzzy inputs and fuzzy risk values relate to the outputs (actions/decisions) based on human expert knowledge.
\end{itemize}

To ensure the correctness and validity of these EDMs, it is essential to verify and validate them during the development process. However, in order to have a reliable EDM model on which many ethical decisions depend, it is fundamental to ensure that the EDM model passes the V\&V criteria.

Now, to be able to automatically map the fEDM model to a fuzzy Petri net, we give the following formal definition of a fEDM.

\paragraph{Definition 1:} A fEDM model (fEDMm) is a 5-tuple: \textit{fEDMm= (fEDMmName, ERFs, ERLs, As/Ds, FERs)} where:
\begin{itemize}
    \item \textit{EDMmName}, is the EDM model name.
    \item \textit{ERFs}, is a set of ethically relevant input facts/factors.
    \item \textit{ERLs}, is a set of possible ethical risk levels (internal state values).
    \item \textit{As/Ds}, is a set of possible actions/decisions to make.
    \item \textit{FERs}, is a set of fuzzy ethical rules.
\end{itemize}
 Here we distinguish between fuzzy ethical rules for risk assessment (FERRs) and fuzzy ethical rules for decision making (FERDs). FERRs are used by the module fERA for the ethical risk assessment. 
\paragraph{Definition 2:} An ethical rule\footnote{An ethical rule is a fuzzy rule with certainty factor.} \textit{R} is a 4-tuple: $R= (Rname, AntS, ConS, CF)$ where:
\begin{itemize}
\item $Rname$, is the name of the rule.
    \item $AntS$, a set of antecedents of the rule. An $AntS$ is a disjunction of conjunctions of antecedents. We denote an antecedent by $A(F)$, where $A$ could be an input (ERF) or an internal state (ERL), and $F$ is a fuzzy set. 
    \item $ConS$, a set of consequents of the rule. We denote a consequent by $C(F)$, where $C$ could be an internal state (ERL), or an output action/decision, and $F$ is a fuzzy set. 
    \item $CF$, is the certainty factor of the rule.
\end{itemize}

According to the above definition of the $EDM$ model, we can distinguish between three types of rules: Type1) $ERFs \xrightarrow{}RLs$, Type2) $RLs \xrightarrow{} AS/DS$, Type3) $ERFs \xrightarrow{} AS/DS$.

\subsection{Fuzzy ERA}
\label{era}
When developing ethical AI systems, the focus should be on identifying the ethical risks associated with both the system and its use. The primary concern is recognizing these risks, rather than delving into the ethical theories that justify why something is considered ethical. Key questions include: What ethical risks are present in the system we are building? How might users employ the system or product in ways that pose ethical risks (deployment considerations)? In response to this, the development team must carefully consider which features to include or exclude in the AI system to effectively mitigate these risks \cite{blackman2022ethical}.
As symbiosis in SAI systems increases, so does the potential for ethical risks, making effective ERA essential. However, due to the inherent fuzziness and incomplete understanding of complex systems, managing this uncertainty is crucial for ensuring ethical behavior and minimizing harm.
We propose a fuzzy logic based system for ERA. The main components of this system for are:
\begin{description}
    \item[Inputs] These are the factors/parameters relevant for the ethical risk calculation.
	\item[Fuzzification] In this stage crisp input values are converted into fuzzy sets, , allowing real-world data (e.g., temperature, speed) to be interpreted in a way that accounts for uncertainty or vagueness. This is done using membership functions that map input values to a degree of membership between 0 and 1. 
	\item[Inference Engine] The inference engine will consults the  \textit{Fuzzy Rules Base} that contains FERRs rules which are a set of "if-then" rules that define the system's behavior. These rules describe how fuzzy inputs relate to the fuzzy output (the ethical risk) based on expert knowledge. The engine will apply these rules to the fuzzified input to derive fuzzy output sets. It determines which rules are relevant based on the degree of membership of the input values.
	\item[Defuzzification] Converting the fuzzy output sets back into crisp values to implement actions or decisions. Common defuzzification methods include \textit{centroid}, \textit{mean of maximum}, and \textit{bisector}, etc.\footnote{\url{https://it.mathworks.com/help/fuzzy/defuzzification-methods.html}} 
 \item[Output] The only Output in our fuzzy system is the ethical risk level. The computed level can be subsequently considered to make the appropriate decision/action to mitigate that risk.
\end{description}

\begin{figure}[htbp!]
  \centering\includegraphics[width=0.9\linewidth]{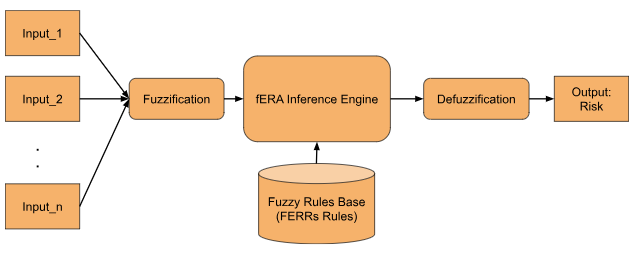}
  \caption{Architecture of the fERA Module}
  \label{fig:fERA}
\end{figure}

Figure \ref{fig:fERA} shows the architecture of the fERA module. The fERA approach will be illustrated using the patient dilemma case study in Section \ref{case}.
\vspace{15pt}


\subsection{Verification Criteria of EDM Models}
\label{ver}
Modeling EDM as a fuzzy rule-base, it may suffer from the structural errors from which any fuzzy rule-base may suffer, among which incompleteness, inconsistency, circularity and redundancy are the most popular \cite{1185845}.
The rule-base verification process consists of detecting and eliminating structural errors.

After mapping the EDM model to a fuzzy Petri net, we will follow the approach of \cite{1185845} for constructing the reachability graph of the resulting Petri net and then use it to detect the different types of structural errors and then eliminate them from our rule base.
The structural errors from which a rule base may suffer are as follows:
\begin{enumerate}
    \item Incompleteness: Incompleteness results from missing rules in a rule base. As an example, let us consider the following set of rules:
    \begin{math}
       r1: \rightarrow P_1 \\
        r2: P_1 \wedge P_3 \rightarrow P_2\\
        r3: P_1 \rightarrow P_4\\
        r4:P_2 \rightarrow\\
    \end{math}
    Rule \( r_1 \) represents a fact, a source transition in the FPN, while rule \( r_4 \) represents a query, a sink transition in the FPN. Rule \( r_2 \) is considered useless because its antecedent \( P_3 \) does not match any part of the consequents in the other rules, making \( P_3 \) a dangling antecedent. Similarly, rule \( r_3 \) is deemed useless because its consequent \( P_4 \) does not match any part of the antecedents in the other rules, rendering \( P_4 \) a dead-end consequent.
    \item Inconsistency: Inconsistent rules lead to conflicts and should be removed from the rule base. This occurs when a set of rules produces contradictory conclusions under certain conditions. An example of inconsistent rules is as follows:
    \begin{math}
       \\r1:  P_1 \wedge P_2 \rightarrow P_3\\
        r2: P_3 \wedge P_4 \rightarrow P_5\\
        r3:  P_1 \wedge P_2 \wedge P_4 \rightarrow \neg P_5\\
    \end{math}
    Rule \( r_3 \) is inconsistent because \( P_1 \) and \( P_2 \) lead to \( P_3 \), and \( P_3 \) and \( P_4 \) lead to \( P_5 \), whereas in \( r_3 \), \( P_1 \) and \( P_2 \) (leading to the same \( P_3 \)) and \( P_4 \) result in \( \neg P_5 \).
    \item Circularity: Circular rules occur when several rules have a circular dependency. This circularity can lead to an infinite reasoning loop and must be resolved. An example of circular rules is as follows:
     \begin{math}
       \\r1:  P_1 \rightarrow P_2\\
        r2: P_2\rightarrow P_3\\
        r3:  P_3 \rightarrow P_1
    \end{math}
    \item Redundancy: Redundant rules are unnecessary in a rule base. They increase the size of the rule base and may lead to additional, useless deductions. An example of redundant rules is as follows:
    \begin{math}
       \\r1:  P_1 \wedge P_3 \rightarrow P_2\\
        r2: P_1 \wedge P_3 \rightarrow P_2\\
         r3:  P_1 \rightarrow P_2\\
         r4:  P_4 \rightarrow P_5\\
        r5: P_4 \rightarrow P_5 \wedge P_6\\
    \end{math}
    Rule \( r_1 \) is redundant to rule \( r_2 \). There are two cases of directly subsumed rules. First, rules \( r_1 \) or \( r_2 \) are subsumed by rule \( r_3 \) because \( r_1 \) or \( r_2 \) have more restrictive conditions than \( r_3 \). Second, rule \( r_4 \) is subsumed by rule \( r_5 \) because \( r_4 \) has a less implied conclusion than \( r_5 \).
\end{enumerate}

\subsection{Validation of EDM Models}
\label{val}
To validate the created model, we will need to construct a validation referent model, together with the domain experts, against which to validate our model. While verification is about dealing with structural errors in the rule base, validation is about dealing with semantic errors. Two types of semantic errors can be identified \cite{4675474}:
\begin{itemize}
    \item Semantic incompleteness: This kind of error occurs if the model does not meet users' requirements and is reflected as missing rules, and/or missing antecedents or consequents in a rule from the users' point of view.
    \item Semantic incorrectness: This type of error occurs if the model produces an output that is different from the expected output for given identical input data in the validation referent. Semantic incorrectness also indicates that the model does not meet the users' needs.
\end{itemize}

\section{A Case Study in the Healthcare Domain}
\label{case}

 To illustrate our proposal for EDM modeling, and then the verification and validation of the created model, we will use the following case study adapted from \cite{AndersonAA06}  which is a common type of ethical dilemma that a care robot may face.\\
\textbf{Patient Dilemma Problem}: \textit{A care robot approaches her competent adult patient to give her her medicine in time and the patient rejects to take it. Should the care robot try again to change the patient’s mind or accept the patient’s decision as final?} 

The dilemma arises because, on the one hand, the care robot may not want to risk diminishing the patient’s autonomy by challenging her decision; on the other hand, the care robot may have concerns about why the patient refuses the treatment. Three of the four Principles/Duties of Biomedical Ethics are likely to be satisfied or violated in dilemmas of this type: the duty of Respect for Autonomy, the duty of Nonmaleficence and the duty of Beneficence.

In this case study, fERA addresses the ethical risk of physical harm to the patient, which works as the basis for decision making by the care robot. In order to evaluate this ethical risk, the care robot can consider different parameters such as the severity of the health condition of the patient, the mental/psychological condition of the patient, physiological indicators of well-being, etc. All these parameters can be considered fuzzy concepts. In order to keep the case study as simple as possible, we consider only two parameters, viz. severity and mental conditions as input. Both inputs are rated on a scale between 0 and 10. The crisp values are then fuzzified, both into 3 sets. For the severity of the health condition, the fuzzy sets are: LOW, MEDIUM, HIGH. For the mental/psychological condition, the fuzzy sets are: BAD, AVERAGE, GOOD.
Starting from these two inputs, once fuzzified, the ERA system calculates the risk level on a scale between 0\% and 100\%. Also for the output there are 3 fuzzy sets: LOW, MEDIUM, HIGH.
The inputs and the output are the antecedents and the consequents, respectively, of the rules employed by the fERA. The fuzzy inference rules used to derive the output of fERA (ethical risk) from the inputs are FERRs.R1, FERRs.R2 and FERRs.R3 in Listing \ref{lst1}.
The initial inputs, in this case severity and mental, are provided by the user (the care robot in this case). These values are then fuzzified using a membership function (MF) \cite{ross2005fuzzy}. In this paper, we choose to use the trapezoidal MF because there is an interval of input crisp values for which the membership degree to the fuzzy set is 100\%.
The fuzzified input is then processed through the above mentioned inference rules
The final output is subsequently defuzzified using centroid method (\cite{Lee1990FuzzyLI}) to find a single crisp value which defines the output of a fuzzy set. This final value provides the level of ethical risk on the life of the patient. fERA was implemented in python using the  Scikit-Fuzzy library for fuzzy logic. We are currently working on the implementation of the whole fEDM system.

Listing \ref{lst1} shows the description of the EDM model for the above case study.

\begin{lstlisting}[mathescape, caption={Formal Description of the EDM Model for the Patient Dilemma},label={lst1}]
EDMm=(PatientEDMm, ERFs, RLs, As/Ds, FERs{FERRs,FERDs})
EDMm.ERFs={Severity, Mental}
EDMm.RLs={Risk}
EDMm.As={Actions/Decisions}
EDMm.FERs={R1,R2,...,R6}
EDMm.ERFs.Severity={Severity, value}
EDMm.ERFs.Severity.value= {low, medium, high}
EDMm.ERFs.Mental={Mental,value}
EDMm.ERFs.Mental.value={good, average, bad}
EDMm.RLs.Risk={RiskLevel,value}
EDMm.RLs.Risk.value={low, medium,high}
EDMm.As.Actions={Action/Decision,value}
EDMm.As.Action.value={accept, tryAgainLater,tryAgainNow}
EDMm.FERs.FERRs.R1=(Rule1, (Severity(low) $\wedge$ Mental(good))$\vee$ (Severity(medium) $\wedge$ Mental(good)) $\vee$ (Severity(low) $\wedge$ Mental(average))$\vee$ (Severity(low) $\wedge$ Mental(bad)), Risk(low),0.80 )

EDMm.FERs.FERRs.R2=(Rule2, (Severity(high) $\wedge$ Mental(good))$\vee$ (Severity(medium) $\wedge$ Mental(average)), Risk(medium),0.70 )

EDMm.FERs.FERRs.R3=(Rule3, (Severity(high) $\wedge$ Mental(average))$\vee$ (Severity(medium) $\wedge$ Mental(bad)) $\vee$ (Severity(high) $\wedge$ Mental(bad)), Risk(high),0.90 )

EDMm.FERs.FERDs.R4=(Rule4, Risk(low), Action(accept),0.80 )
EDMm.FERs.FERDs.R5=(Rule5, Risk(high), Action(try_again_now),0.90 )
EDMm.FERs.FERDs.R6=(Rule6, Risk(medium), Action(try_again_later),0.70 )
\end{lstlisting}

\subsection{Verification Process}
The steps we follow in the verification process of the rule-base are:
\begin{itemize}
    \item \textbf{Step1}: rules normalization: Apply normalization rules introduced in \cite{1185845} to change any rule to a rule of the following form:
    
    $P_1 \wedge P_2 \wedge ... P_{j-1} \rightarrow P_j$
    \item \textbf{Step2}: transformation of the rule-base to FPN: The antecedents of a rule are mapped to input places of FPN, consequents of a rule to output places of FPN, and a rule name to a transition of FPN.
    \item \textbf{Step3}: generating the FPN reachability graph: Herein, we adopt the algorithm to generate the reachability graph in \cite{814327}. We generate a reachability graph like a coverability tree in Petri net theory to check for structural errors. Each node in the reachability graph is a marking of the static analytic Petri net model. Each directed edge represents a transition firing and connects one node to the other.
    \item \textbf{Step4}: looking for errors by checking the reachability graph: we verify the incompleteness, inconsistency, circularity, and redundancy of the rule bases by checking the reachability graph \cite{814327}.
\end{itemize}

By applying rule normalization, Rule1 will result in four rules, Rule2 in two rules, Rule3 in three rules, Rules 4,5, and 6 in one rule each. The total number of resulting rules after normalization is 12, see the listing \ref{lst2}.

\begin{lstlisting}[mathescape, caption={Normalized Rules},label={lst2}]
(Severity(low) $\wedge$ Mental(good))$\rightarrow$ Risk(low)
(Severity(medium) $\wedge$ Mental(good))$\rightarrow$ Risk(low)
(Severity(low) $\wedge$ Mental(average))$\rightarrow$ Risk(low)
(Severity(low) $\wedge$ Mental(bad)) $\rightarrow$ Risk(low)
(Severity(high) $\wedge$ Mental(good))$\rightarrow$ Risk(medium)
(Severity(medium) $\wedge$ Mental(average))$\rightarrow$ Risk(medium)
(Severity(high) $\wedge$ Mental(average)) $\rightarrow$ Risk(high)
(Severity(medium) $\wedge$ Mental(bad)) $\rightarrow$ Risk(high)
(Severity(high) $\wedge$ Mental(bad)) $\rightarrow$ Risk(high)
Risk(low) $\rightarrow$ Action(accept)
Risk(high) $\rightarrow$ Action(try_again_now)
Risk(medium) $\rightarrow$ Action(try_again_later)
\end{lstlisting}

\begin{figure}[ht]
  \centering
  \includegraphics[width=0.9\linewidth]{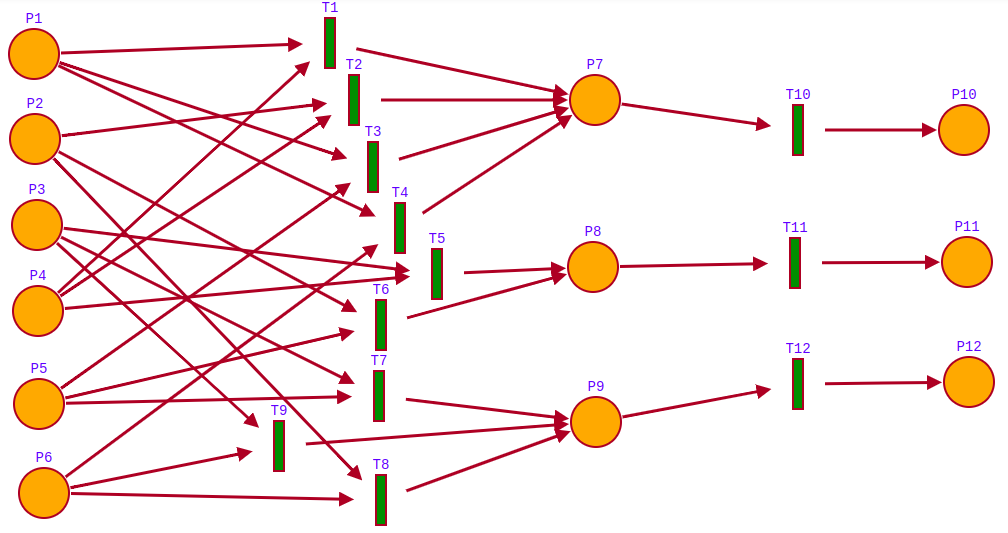}
  \caption{Fuzzy Petri Net of the Patient EDM Model}
  \label{fig:petri}
\end{figure}

The transformation of this rule base to FPN will result in the FPN shown in figure \ref{fig:petri}. We have twelve places and twelve transitions. The transitions $t_1 ... t_{12}$ corresponds to the rules in listing \ref{lst2} in the same order. While $P_1 ... P_{12}$ corresponds to: severity(low), severity(medium), severity(high), mental(good), mental(average), mental(bad), risk(low), risk(medium), risk(high), action(accept), action(try\_again\_later), action(try\_again\_now) respectively.

To check for structural errors, we need to generate the reachability graph of the resulting FPN. Each node in the reachability graph is a marking of the FPN model. Each directed edge represents transition firing and connect one node to the other. For generating the reachability graph we adopt the algorithm presented in \cite{1185845}.
After generating the reachability graph, the structural errors including incompleteness, inconsistency, redundancy and circularity are checked and eliminated.

\begin{figure}[htbp!]
  \centering
  \includegraphics[width=0.6\linewidth, height=4cm]{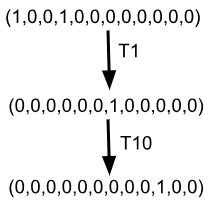}
  \caption{Reachability Graph (version 1)}
  \label{fig:reach}
\end{figure}

The corresponding reachability graph for the above FPN is shown in figure \ref{fig:reach}. In the initial marking, places $P_1$ and $P_4$ are regarded as true antecedents, and initially filled (set to $1$) for this reason, the rest of places are set to $0$, that's why in the first node there are two one's. This causes $t_1$ to fire in the first step. After firing this transition, in the second step, place $P_7$ is filled and the corresponding value in the node vector is set to $1$. On the final step, by firing $t_{10}$ (the enabled transition), place $P_{10}$ will be filled up.

All the places and transitions exist so there are no incompleteness errors. $P_1$, $P_2$, and $P_3$ are different states of one property (severity). Similarly,  $P_4, P_5, P_6$ are different states of one property (mental), $P_7, P_8, P_9$ are different states of one property (risk), $P_{10}, P_{11}, P_{12}$ are different states of one property (action), Their simultaneous existence may refer to some concept of inconsistency in the rule base. The reachability graph has no loops means no circularity errors and finally having no transitions underlined, speaks for non-redundancy.

 \vspace{10pt}

\subsection{Validation Process}
To validate the created EDM model, we will need to construct a validation referent. The validation referent provides a standard against which to compare our EDM models to determine the validity of these models for our needs. Validation referent gives the key information that a model must represent. Validation referent should be developed by experienced domain experts and expressed in the same language that is used to create the EDM model.

To build the validation referent, validation  evidences are collected from domain experts and recorded in the validation referent. There exists three types of validation evidences (rules): Type1) $ERFs \xrightarrow{}RLs$, Type2) $RLs \xrightarrow{} AS/DS$, Type3) $ERFs \xrightarrow{} AS/DS$.
To ensure the validity of the EDM models, these types of rules above should be deduced from the EDM models. Validation will be carried out in two stages: first) static validation, and Second) dynamic validation.

Listing \ref{lst3} shows the validation referent for our case study. $LTconsequences$ refer to the long term consequences that might result from non taking the medication by the patient. $RR1$ and $RR2$ are the reasoning rules, with reference values given by the domain expert, that will serve for the dynamic validation. $\alpha$ is used to refer to the degree of truth of a proposition.

\begin{lstlisting}[mathescape, caption={Formal Description of the Referent Model for the Patient EDM Model},label={lst3}]
EDMmr=(PatientEDMmr, ERFsr, RLsr, Asr/Dsr, FERsr{FERRsr, FERDsr})
EDMmr.ERFsr={Severity, Mental, LTconsequences}
EDMmr.RLsr={Risk}
EDMmr.Asr={Actions/Decisions}
EDMmr.Rsr={R1,R2,...,R6}
EDMmr.ERFsr.Severity={Severity, value}
EDMmr.ERFsr.Severity.value= {low, medium, high}
EDMmr.ERFsr.Mental={Mental,value}
EDMmr.ERFsr.Mental.value={good, average, bad}
EDMmr.ERFsr.LTconsequences={LTconsequences, value}
EDMmr.ERFsr.Ltconsequences.value= {low, medium, high}
EDMmr.RLsr.Risk={RiskLevel,value}
EDMmr.RLsr.Risk.value={low, medium,high}
EDMmr.Asr.Actions={Action/Decision,value}
EDMmr.Asr.Action.value={accept, tryAgainLater,tryAgainNow}
EDMmr.FERsr.FERRsr.R1r=(Rule1, ((Severity(low) $\wedge$ Mental(good))$\vee$ ((Severity(medium) $\wedge$ Mental(good)) $\vee$ ((Severity(low) $\wedge$ Mental(average))$\vee$ ((Severity(low) $\wedge$ Mental(bad)), Risk(low),0.80 )
EDMmr.FERsr.FERRsr.R2r=(Rule2, ((Severity(high) $\wedge$ Mental(good))$\vee$ ((Severity(medium) $\wedge$ Mental(average)), Risk(medium),0.70 )
EDMmr.FERsr.FERRsr.R3r=(Rule3, ((Severity(high) $\wedge$ Mental(average))$\vee$ ((Severity(medium) $\wedge$ Mental(bad)) $\vee$ ((Severity(high) $\wedge$ Mental(bad)), Risk(high),0.90 )


EDMmr.FERsr.FERDsr.R4r=(Rule5, Risk(high), Action(tryAgainNow),0.95 )
EDMmr.FERsr.FERDsr.R5r=(Rule4, Risk(low) $\wedge$ LTconsequences(low), Action(accept),0.70 )

EDMmr.FERsr.FERDsr.R6r=(Rule6, Risk(medium) $\vee$ ((Risk(low) $\wedge$ LTconsequences(medium)) $\vee$ ((Risk(low) $\wedge$ LTconsequences(high)), Action(tryAgainLater),0.80 )

RR1: $\alpha$(Severity(high))$=0.9 \wedge \quad \alpha$(Mental(bad))$=0.8$) $\rightarrow \alpha$(Risk(high)) $> 0.7$ 
RR2: $\alpha$(Severity(medium))$=0.9 \wedge \quad \alpha$(Mental(good))$=0.8$) $\wedge \quad \alpha$(LTconsequences(medium))$=0.8$)$\rightarrow \alpha$(Action(tryAgainLater)) $> 0.7$ 
\end{lstlisting}

\paragraph{Static Validation (Semantic Completeness):}
In this stage, we are going to search the FPN that represents the EDM model and compare the searching results with the validation referent to determine if our model is semantically incomplete. For conducting static validation, we are going to follow this procedure \cite{4675474}:
\begin{itemize}
    \item Step1: searching the places of the FPN that represents the EDM model.
    \item Step2: recording the properties that correspond to each place into the corresponding $ERFs, RLs, As/Ds$. 
    \item Step3: compare the searched $ERFs, RLs, As/Ds$ with those of the validation referent model $ERFsr, RLsr, Asr/Dsr$.
    \item Step4: if the number of the searched $ERFs$ is less than those of the referent model then the model may misses antecedents. If the number of the searched $As/Ds$ is less than those of the referent model then the model may misses consequents. If the number of the searched $RLs$ is less than those of the referent model then the model may misses antecedents or consequents. If an expected rule doesnot appear in the FPN, then the model might misses rules.
\end{itemize}

By applying the above steps on our $PatientEDM$ model, we found that: the property $LTconsequences$ which is an ethically relevant fact does not appear in the FPN of our $PatientEDM$ model, and also the rule $R6r$ is missing. Therefore, we can conclude that our model is semantically incomplete with respect to the referent model. By adding these missing components we render our model semantically complete with respect to the validation referent model presented in listing \ref{lst3} and discussed with the domain experts.
\begin{figure}[ht]
  \centering
  \includegraphics[width=0.9\linewidth]{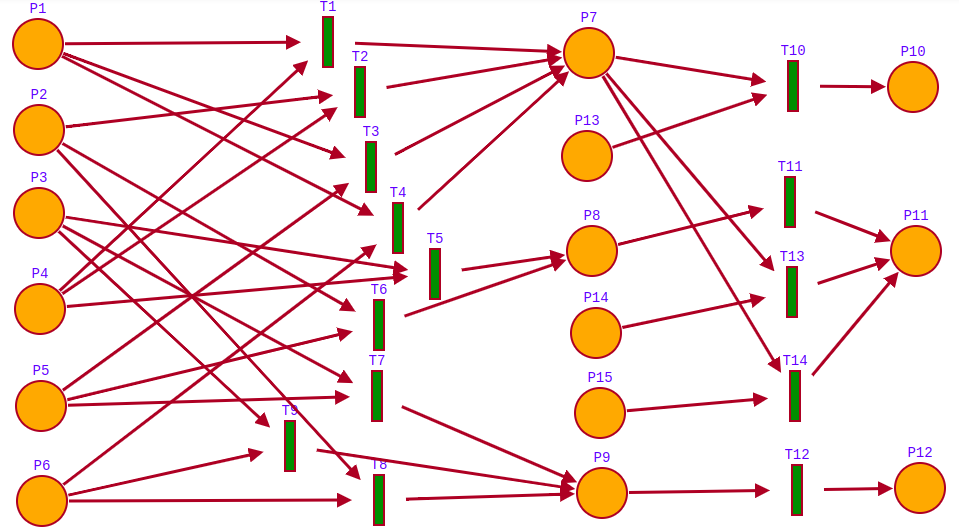}
  \caption{Fuzzy Petri Net of the Referent Patient EDM Model}
  \label{fig:petri_r}
\end{figure}
\vspace{15pt}

\paragraph{Dynamic Validation (Semantic Correctness):}
This step involves checking for semantic incorrectness through running and reasoning over the FPN. We need to reason the FPN for given inputs and compare the results to their counterparts in the referent model to check if there exists any semantic incorrectness.

The uncertainty reasoning of the three types of
rules with certainty factors (see section \ref{petri}) can be summarized as follows \cite{60794}:

\begin{itemize}
	\item Type 1: $R_{i}(\beta_i): P_1 (\alpha_1) \wedge P_2 (\alpha_2)\wedge ... \wedge P_{j-1} (\alpha_{j-1})\rightarrow P_j (\alpha_j)\wedge P_{j+1} (\alpha_{j+1})\wedge ... \wedge P_k (\alpha_k)$ \\
 $\alpha_j = \alpha{j+1} = ... = \alpha_k = min\{\alpha_1, \alpha_2, ... \alpha_{j-1}\}*\beta_i .$
 
	\item Type 2: $R_1(\beta_1) : P_j(\alpha_j) \rightarrow P_1(\alpha_1); \quad
R_2(\beta_2) : P_j(\alpha_j) \rightarrow P_2(\alpha_2); \quad \dots \quad;
R_j(\beta_{j-1}) : P_j(\alpha_j) \rightarrow P_{j-1}(\alpha_{j-1}).$ \\
$\alpha_1 = \alpha_j * \beta_1 . \alpha_2 = \alpha_j * \beta_2 ... \alpha_{j-1} = \alpha_j * \beta_{j-1} .$

	\item Type 3: $R_{i}(\beta_i): (P_1 (\alpha_1)\vee P_2 (\alpha_2)\vee ... \vee P_{j-1} (\alpha_{j-1}))\rightarrow P_j (\alpha_j)$ \\
 $\alpha_j = max \{\alpha_1 , \alpha_2 , ... , \alpha_{j-1} \} * \beta_i .$
\end{itemize}

The first step in this phase is to map the revised $PatientEDM$ model after static validation into FPN. Figure\ref{fig:petri_r} shows the FPN of the revised model and Listing\ref{lst4} shows its specifications.
Now, we can reason over the FPN applying the reasoning rules and comparing the results to the reasoning references ($RR1$ and $RR2$ in our case) from the validation referent.

\begin{lstlisting}[mathescape, caption={Specification of the FPN}, label={lst4},  breaklines=true, breakatwhitespace=true]

FPN = (P, T, D, I, O, $\mu$, $\alpha$, $\beta$);
P = {$p_1, p_2, p_3, p_4, p_5, p_6, p_7, p_8, p_9, p_{10}, p_{11}, p_{12}, p_{13}, p_{14}, p_{15}$};
T = {$t_1, t_2, t_3, t_4, t_5, t_6, t_7, t_8, t_9, t_{10}, t_{11}, t_{12}, t_{13}, t_{14}$};
D = {Severity(low), Severity(medium), Severity(high), Mental(good), Mental(average), Mental(bad), Risk(low), Risk(medium), Risk(high), Action(accept), Action(tryAgainLater), Action(TryAgainNow), LTconsequences(low), LTconsequences(medium), LTconsequences(high)};
$\mu$ = (0.80, 0.70, 0.90, 0.95, 0.70, 0.80);
$p_1$ = Severity(low), $p_2$ = Severity(medium), 
$p_3$ = Severity(high), $p_4$ = Mental(good), 
$p_5$ = Mental(average), $p_6$ = Mental(bad), 
$p_7$ = Risk(low), $p_8$ = Risk(medium), 
$p_9$ = Risk(high), $p_{10}$ = Action (accept),
$p_{11}$= Action(tryAgainLater), 
$p_{12}$ = Action(TryAgainNow), 
$p_{13}$ = LTconsequences(low), 
$p_{14}$ = LTconsequences(medium), 
$p_{15}$ = LTconsequences(high).

\end{lstlisting}

Let us see $RR1$ in the validation referent in Listing \ref{lst3}. Using the FPN to reason, we can obtain the following reasoning results: $\alpha (Risk(high)) = min (0.9,0.8)*0.9 = 0.8*0.9 = 0.72 > 0.70$. therefore, the validation rule of $RR1$ has passed through the validation.
Let us see $RR2$ in the validation referent in Listing \ref{lst3}. Using the FPN to reason, we can obtain the following reasoning results: $\alpha (Action(TryAgainLater)) = min ( min (0.9,0.8)*0.8), 0.8) = min (0.64, 0.70) = 0.64 < 0.70$. therefore, the validation rule of $RR2$ has not passed through the validation, this means in $PatientEDM$ model there are some semantic errors that needs to be corrected.

\section{Discussion and Conclusion}
\label{con}

In this work, we proposed a novel fuzzy rule-based representation for modeling EDM systems.
In our modeling, the EDM is mainly based on ERA. The calculated level of ethical risk will guide the final decision for the sake of ethical risk mitigation.
Our formal representation method of EDM models can be used for creating EDM models in any domain.

In our model, it is not about delineating what is morally permissible from a deontological point of view, nor what is morally permissible from a consequentialist point of view, neither from a virtuous stand point. It is about executing the task at hand in a 'morally appropriate' way. Means for us, choosing the best action/decision (based on ERA) to mitigate the possible ethical risk that the stakeholders of the SAI system might get posed to.
However, in our definition, the EDM model based on ERA, indirectly adheres to the ethical framework of interest, respecting ethical norms and values of the considered domain. That is because the ethical risk might include, for example, the risk of violating values or norms like privacy, confidentiality, fairness, etc.

In this work, we considered a simple case study adapted from the literature of machine ethics, to illustrate our proposal for EDM modeling. In the future, we are planning to consider more complex case studies and conduct more real world experiments. 

The proposed formal representation of EDM models allows us to verify the created models using formal verification methods (see section \ref{case}). Furthermore, it allows us to construct validation referents against which to validate the created models. Validation referents can be based on non-experts (using folk morality), domain experts (the most common source), or legal standards. Each approach has its own limitations depending on the specific domain.
Validation referents construction is subject to our future work. Furthermore, we are also investigating the existing FPN based tools for automatic verification and validation of our fEDM models.
Finally, there is a need for more systematic evaluations when ethical machines are created to be able to rate and compare systems. To this end, this work contributes to the achievement of this objective.

\section*{Acknowledgments}
  This work was partially supported by the project FAIR - Future AI Research (PE00000013), under the NRRP MUR program funded by the NextGenerationEU.


\bibliography{fuzzy}


\end{document}